\documentclass{article}

\usepackage{graphicx} 
\usepackage{subfigure} 

\usepackage{natbib}

\usepackage{algorithm}
\usepackage{algorithmic}

\usepackage{amsmath}
\usepackage{amssymb}
\usepackage{amsfonts}
\usepackage{amsthm}

\newcommand{\reals}{\mathbb{R}}
\newcommand{\bx}{\mathbf{x}}
\newcommand{\bc}{\mathbf{c}}

\title{The Artificial Regression Market}
\author{Nathan Lay (nlay@fsu.edu), Adrian Barbu (abarbu@stat.fsu.edu)}

\begin{document} 
\maketitle
\date{}






\begin{abstract} 
The Artificial Prediction Market is a recent machine learning technique for multi-class classification, inspired from the financial markets. It involves a number of trained market participants that bet on the possible outcomes and are rewarded if they predict correctly.
This paper generalizes the scope of the Artificial Prediction Markets to regression, where there are uncountably  many possible outcomes and the error is usually the MSE. For that, we introduce the reward kernel that rewards each participant based on its prediction error and we derive the price equations. Using two reward kernels we obtain two different learning rules, one of which is approximated using Hermite-Gauss quadrature.
The market setting makes it easy to aggregate specialized regressors that only predict when an observation falls into their specialization domain. Experiments show that regression markets based on the two learning rules outperform Random Forest Regression on many UCI datasets and are rarely outperformed.

\end{abstract} 

\section{Introduction}
\noindent Prediction markets are forums of trade where contracts on the outcomes of future events are bought and sold.  Each contract is a wager that yields payment if its corresponding outcome occurs. Each market participant has an incentive to profit and therefore an incentive to predict accurately. The trading prices of contracts are determined by supply and demand. Highly demanded contracts are more expensive and represent an overall confidence that a corresponding outcome will be realized. On the other hand, less demanded contracts are less expensive and represent an overall lack of confidence that a corresponding outcome will be realized. These trading prices can be interpreted as the market's prediction of the outcome. Studies have shown that the trading prices even estimate the true probability of the outcome~\cite{manski2006interpreting}. Prediction markets have found use in predicting elections, decision making in both government and business realms, and even sporting events~\cite{arrow_promise_2008}.  Their reported accuracy and success motivated the development of the Artificial Prediction Market~\cite{lay2009,lay2010apm,market2011} that attempts to mimic a real prediction market in a machine learning setting. The Artificial Prediction  Market has empirically proven to be a competitive classifier aggregation technique and motivates further investigation. 
It was proved in \citet{market2011} that the Artificial Prediction Market learns by constrained Maximum Likelihood.

In this paper we generalize the Artificial Prediction Market to regression. While the objective of classification is to predict a label from a finite set of labels, the objective of regression is to predict a real value response. We develop a mathematical analog of the Artificial Prediction Market, the Regression Market, to deal with real values, or uncountably many ``labels''. Regression markets are unusual in that contracts are no longer discrete and finite. Each contract corresponds to a real value prediction and consequently there are uncountably many such contracts for trade. While in classification a contract that has not predicted the correct outcome does not win anything, for regression we introduce the reward kernel that rewards contracts based on the distance to the ground truth value.

We further show experiments on UCI~\cite{uci2010} and LIAAD~\cite{liaad2010} data sets that demonstrate that the Regression Market is a viable technique for aggregating regressors, and also works very well with specialized regressors that only predict outcomes for certain instances and not for other.

\section{Related Work}
To the best of our knowledge, there has been no other work on solving regression tasks with machine learning models of prediction markets. Related work can be found for classification in~\citet{lay2010apm,market2011} where \textit{Artificial Prediction Markets} were developed for classification using betting functions and an equilibrium based on conservation of budget sum.

 Another model can be found in~\cite{storkey2011} where \textit{machine learning markets} are instead derived from utility functions. 

In~\cite{jenny2010} the authors find a connection between no-regret learning and prediction markets.

\section{Overview of the Artificial Prediction Markets}

\noindent In ~\citet{lay2010apm}, the classification market is defined by a \textit{betting function} $\phi^k(\bx,\bc)$ that describes the proportion of the budget $\beta$ to allot for label $k$ for a given instance $\bx$ and trading prices for all labels $\bc$. The equilibrium price $\bc$ is defined such that the for any label, the sum of profits equaled the sum of losses
\[
	\sum_{m=1}^M \beta_m \frac{\phi_m^y(\bx,\bc)}{c_y} = \sum_{m=1}^M \beta_m \sum_{k=1}^K \phi_m^k(\bx,\bc) \quad y = 1, 2, \hdots, K
\]
This equilibrium system corresponds to the update rule for the classification market
\[
	\beta_m \leftarrow \beta_m - \beta_m \sum_{k=1}^K \phi_m^k(\bx,\bc) + \beta_m \frac{\phi_m^y(\bx,\bc)}{c_y}
\]
for $m=1,2,\hdots,M$. This is the profit. With a little reworking, the above equilibrium is equivalent to solving the following fixed point problem
\[
	c_k = \sum_{m=1}^M \beta_m \phi_m^k(\bx,\bc) \quad k = 1, 2, \hdots, K
\]
The trading price $\bc$ is considered to be an estimate of the conditional mass. In fact, ~\citet{market2011} demonstrates that the classification market maximizes log likelihood.

\section{Regression Markets}

\noindent The extension of prediction markets to the regression problem proves to be counterintuitive. In classification, the goal is to predict the one correct label for a given instance. What can be said about regression? Assume, for the time being that the classification market framework generalizes. For the sake of consistency with probability notation $\phi(y|\bx,c)$ will denote a betting functional that allots a proportion of the budget for response $y \in \reals$. This implies that
\begin{equation}
	0 \leq \int_Y \phi(y|\bx,c) dy \leq 1
\end{equation}
since no participant may bet more than the whole of their budget in this market. A curious consequence of this constraint is that it is possible for $\phi(y|\bx,c) > 1$ for some $y$. Likewise, the trading prices for $y$ are denoted as the price function $c(y|\bx)$. The trading price is a conditional density on the possible responses $y$. The prediction can be computed from, for example, expectation
\begin{equation}
	y = \int_Y t c(t|\bx) dt
\end{equation}
However, the price function can also model ambiguous responses. For example, points along a circle could result in a bimodal price function.

\noindent The equilibrium price function $c(y|\bx)$ receives similar treatment as the classification market. The objective is to find a $c(y|\bx)$ that gives conservation of budget. The ambiguity of the \textit{correct} label mentioned above is resolved by introducing a \textit{reward kernel} $K(t;y)$. The reward kernel is a density with a single mode centered about the ground truth $y$. The winnings are subsequently defined as
\begin{equation}
\text{winnings} = \int_Y K(t;y) \frac{\phi(t|\bx,c)}{c(t|\bx)} dt
\end{equation}
and bears similarity to the winnings in the classification market. This has the effect of partially rewarding participants for nearby predictions. Likewise, the total expenditures for contracts are given as
\begin{equation}
\text{bet} = \int_Y \phi(t|\bx,c) dt
\end{equation}
Analogous to the classification market, the equilibrium price function $c(y|\bx)$ is defined such that gains match losses
\begin{equation}
\sum_{m=1}^M \beta_m \int_Y K(t;y) \frac{\phi_m(t|\bx,c)}{c(t|\bx)} dt = \sum_{m=1}^M \beta_m \int_Y \phi_m(t|\bx,c) dt
\end{equation}

\subsection{Constant Market for Regression}
\noindent For simplicity and the reported empirical performance of the \textit{constant classification market}, the remainder of this paper assumes $\phi(y|\bx,c) = h(y|\bx)$ where $h(y|\bx)$ is a conditional density with mean $f(\bx)$. Here $f(\bx)$ is a regressor. This defines the \textit{constant market} for regression with
\begin{align}
	c(y|\bx) &= \sum_{m=1}^M \beta_m h_m(y|\bx) \\
	y &= \int_Y t c(t|\bx) dt =  \sum_{m=1}^M \beta_m f_m(\bx)
\end{align}
The update rule is similar to that of the classification market in exception to the additional \textit{reward kernel}
\begin{equation}
	\beta_m \leftarrow \beta_m + \eta \beta_m \left ( \int_Y K(t;y) \frac{h_m(t|\bx)}{c(t|\bx)} dt - 1 \right )
\end{equation}
where $\eta$ is the learning rate and also serves to prevent instanaeous bankruptcy (i.e. $\beta = 0$). The choice of $K(t;y)$ gives different update rules. We examine $K(t;y) = \delta(t-y)$ where $\delta(t)$ is the Dirac delta function and $K(t;y) = \frac{1}{\sqrt{2 \pi} \sigma} e^{\frac{-(t-y)^2}{2 \sigma^2}}$

\subsection{Delta Updates}
\noindent When $K(t;y) = \delta(t-y)$ this gives an analogous update rule as the \textit{classification market}
\begin{equation}
	\beta_m \leftarrow \beta_m + \eta \beta_m \left ( \frac{h_m(y|\bx)}{c(y|\bx)} - 1 \right )
	\label{eq:deltaupdate}
\end{equation}
Even though this reward kernel is exacting, it will be shown empirically to work relatively well.

\subsection{Gaussian Updates}
\noindent When $K(t;y) = \frac{1}{\sqrt{2 \pi} \sigma} e^{\frac{-(t-y)^2}{2 \sigma^2}}$, this gives an update involving an integral
\begin{equation}
	\beta_m \leftarrow \beta_m + \eta \beta_m \left ( \int_{-\infty}^{\infty} \frac{1}{\sqrt{2 \pi} \sigma} e^{-\frac{-(t-y)^2}{2 \sigma^2}} \frac{h_m(t|\bx)}{c(t|\bx)} dt - 1 \right )
	\label{eq:hermiteupdate}
\end{equation}
One way to approximate this integral is with Hermite-Gauss quadrature~\cite{press2007numerical}. A change of variables is required to apply the quadrature rule
\begin{align}
	& \int_{-\infty}^{\infty} \frac{1}{\sqrt{2 \pi} \sigma} e^{-\frac{-(t-y)^2}{2 \sigma^2}} \frac{h_m(t|\bx)}{c(t|\bx)} dt \\
	& = \frac{1}{\sqrt{\pi}} \int_{-\infty}^{\infty} e^{-t^2} \frac{h_m(y + \sqrt{2} \sigma t|\bx)}{c(y + \sqrt{2} \sigma t|\bx)} dt \\
	& \approx \frac{1}{\sqrt{\pi}} \sum_{i=1}^n \omega_i \frac{h_m(y + \sqrt{2} \sigma t_i|\bx)}{c(y + \sqrt{2} \sigma t_i|\bx)}
\end{align}
where $\omega_i,\ t_i$ are the $n$-point Hermite-Gauss weights and nodal points. 

\noindent Intuitively, the choice of $\sigma$ should reflect the noise variance of the training data (assuming Gaussian noise). If $\sigma$ is too small, the market is more prone to overfitting. This $\sigma$ can be chosen with cross validation by discretizing $\alpha \in (0,1]$ and trying $\sigma = \alpha \sqrt{\frac{1}{N} \sum_{n=1}^N y_n^2}$ (assuming the noise has mean $0$).

\subsection{Specialized Regression Markets}
Introduced in~\citet{lay2010apm}, specialized markets are markets with participants which have local support in the feature space. This type of participant is assumed to perform relatively well in its domain. An example of a specialized market is a market with random tree leaves as participants. These types of markets have been demonstrated to be competitive with random forest. The specialized regression market of tree leaves is similar except that leaves are Gaussian instead of histograms. Each regression tree stores the sample mean $\bar{y}$ and variance $\sigma^2$ of instances that fall in each leaf.

\section{Results}
\begin{table*}[htb]
\label{tbl:breiman}
\caption{Table of MSE for forests and markets on UCI and LIAAD data sets. The $F$ column is the number of inputs, $Y$ is the range of regression, RFB is Breiman's reported error, RF is our forest implementation, DM is the Market with delta updates, and GM is the Market with Gaussian updates. Bullets/daggers represent pairwise significantly better/worse than RF while +/-- represent significantly better/worse than RFB.}
\tiny
\centering
\begin{tabular}{|c|c|c|c|c|c|c|c|c|}
\hline
Data & $N_{\text{train}}$ & $N_{\text{test}}$ & $F$ & $Y$ & RFB & RF & DM & GM \\
\hline
abalone & 4177 & -- & 8 & $[1.00,29.00]$ & 4.600 & 4.571 & 4.571 & 4.571 \\
friedman1 & 200 & 2000 & 10 & $[4.30,26.03]$ & 5.700 & 4.343+ & 4.335\textbullet+ & 4.193\textbullet+ \\
friedman2 & 200 & 2000 & 4 & $[-167.99,1633.87]$ & 19600.0 & 19431.852 & 19232.482\textbullet & 18369.546\textbullet+ \\
friedman3 & 200 & 2000 & 4 & $[0.13,1.73]$ & 0.022 & 0.028-- & 0.028\textbullet-- & 0.026\textbullet-- \\
housing & 506 & -- & 13 & $[5.00,50.00]$ & 10.200 & 10.471 & 10.130\textbullet & 10.128\textbullet \\
ozone & 330 & -- & 8 & $[1.00,38.00]$ & 16.300 & 16.916 & 16.925 & 16.917 \\
servo & 167 & -- & 4 & $[0.13,7.10]$ & 0.246 & 0.336 & 0.295 & 0.322 \\
ailerons & 7154 & 6596 & 40 & $[-0.00,-0.00]$ & -- & 2.814e-008 & 2.814e-008\textbullet & 2.814e-008\textbullet \\
auto-mpg & 392 & -- & 7 & $[9.00,46.60]$ & -- & 6.469 & 6.444 & 6.405\textbullet \\
auto-price & 159 & -- & 15 & $[5118.00,35056.00]$ & -- &3823550.43 & 3723413.430 & 3815863.98 \\
bank & 4500 & 3693 & 32 & $[0.00,0.67]$ & -- & 7.238e-003 & 7.212e-003\textbullet & 7.210e-003\textbullet \\
breast cancer & 194 & -- & 32 & $[1.00,125.00]$ & -- & 1112.270 & 1112.509 & 1108.325 \\
cartexample & 40768 & -- & 10 & $[-12.69,12.20]$ & -- & 1.233 & 1.233\dag & 1.232\textbullet \\
computeractivity & 8192 & -- & 21 & $[0.00,99.00]$ & -- & 5.414 & 5.398\textbullet & 5.414\dag \\
diabetes & 43 & -- & 2 & $[3.00,6.60]$ & -- & 0.415 & 0.426\dag & 0.415 \\
elevators & 8752 & 7847 & 18 & $[0.01,0.08]$ & -- & 9.319e-006 & 9.288e-006\textbullet & 9.225e-006\textbullet \\
forestfires & 517 & -- & 12 & $[0.00,1090.84]$ & -- & 5834.819 & 5844.493\dag & 5680.131\textbullet \\
kinematics & 8192 & -- & 8 & $[0.04,1.46]$ & -- & 0.013 & 0.013\textbullet & 0.013\textbullet \\
machine & 209 & -- & 6 & $[6.00,1150.00]$ & -- & 3154.521 & 2991.798\textbullet & 3042.336 \\
poletelecomm & 5000 & 10000 & 48 & $[0.00,100.00]$ & -- & 29.813 & 28.855\textbullet & 29.863\dag \\
pumadyn & 4499 & 3693 & 32 & $[-0.09,0.09]$ & -- & 9.237e-005 & 8.917e-005\textbullet & 8.888e-005\textbullet \\
pyrimidines & 74 & -- & 27 & $[0.10,0.90]$ & -- & 0.013 & 0.013 & 0.012 \\
triazines & 186 & -- & 60 & $[0.10,0.90]$ & -- & 0.015 & 0.015 & 0.015 \\
\hline
\end{tabular}

\end{table*}

\noindent We performed two types of experiments with both updates~\eqref{eq:deltaupdate},~\eqref{eq:hermiteupdate} and compared with Breiman's original regression results~\cite{breiman_random_2001} as well as additional data sets from UCI and LIAAD~\cite{liaad2010}. To be consistent with Breiman, nearly all experiments were conducted over 100 random splits where each split randomly sets aside 10\% of the data set for testing. For abalone, only 10 random splits with 25\% of the data set aside for testing were considered. Data sets with provided test sets were not randomly split. Instead, the forest and markets were trained 100 times on the entire training set and tested on the provided test set. These results vary due to the randomness of the regression forest.

\noindent All experiments were run on Windows 7 with 8GB of RAM and Core i7-2630QM process (max 2.9GHz, 6MB L3 cache). On each training set 100 regression trees were trained. Each regression tree node considered 25 randomized features, each a linear combination of 2 random inputs. Each coefficient of the linear combination was uniformly picked from $[-1,1]$. In our implementation, 1000 of these random features were generated in advance rather than at each node. The split criteria for each node is based on the weighted sample variance. The rule ``don't split if the sample size is $< 5$'' was enforced. Additionally, our implementation treats categoricals as numeric inputs which differs from Breiman's implementation. However, most data sets are comprised of numeric inputs.

\noindent Both market types were trained and evaluated over 50 epochs. Each epoch is one complete pass through the training set. The reported errors are those that minimize the MSE of the test set over the 50 epochs (averaged over the 100 runs).
\begin{equation}
	\text{MSE} = \frac{1}{N} \sum_{n=1}^N (f(\bx_n)-y_n)^2
\end{equation}
The learning rate $\eta = \frac{10}{N_{\text{train}}}$ was used as in~\citet{market2011}. On the first run (random split or full training set), the parameter $\sigma$  for the Gaussian Market reward kernel was estimated using 2-fold cross validation on the training set. This $\sigma$ remained constant for the other $99$ runs ($9$ runs for abalone). The Gaussian market used 5-point Hermite Gauss quadrature. The prediction for $y$ was computed with expectation
\begin{equation}
	y = \int_Y t c(t|\bx) dt = \sum_{m=1}^M \beta_m f_m(\bx)
\end{equation}

\noindent In every result, significance is measured with significance level $\alpha = 0.01$ in two ways: pairwise t-test~\cite{demsar2006statistical} and t-test on the means. The pairwise t-test was used to compare the 100 market runs with the 100 forest runs while the t-test on the means were compared with Breiman's reported results.


\subsection{Comparison with Random Forest Regression}
\noindent The first experiment considers aggregation of tree leaves of forests with fully grown trees on UCI and LIAAD data sets. The results of seven of the data sets are compared with Breiman's reported results. The missing data set Robot Arm is private.

\noindent From~\ref{tbl:breiman} our RF doesn't perform identically with RFB. This can be attributed to the synthetic nature of some data sets such as friedman1, friedman2, and friedman3 and/or the fact that our implementation of regression forest does not treat categorical inputs the same way. Of the Breiman comparisons, only GM is legitimately significantly better than Breiman's results for friedman2. Out of all the data sets, DM is significantly better than RF for 12 data sets (in a pairwise sense) while GM is only significantly better than RF for 11 data sets. However, DM is significantly worse than RF for 3 data sets while GM is only significantly worse on 2 data sets. The significantly worse results can be attributed to overfitting and/or poorly tuned reward kernel in the case of GM.
\begin{table*}[htb]
\label{tbl:shallow}
\caption{Table of MSE for depth 5 forests and markets on UCI and LIAAD data sets. The $F$ column is the number of inputs, $Y$ is the range of regression, RFB is Breiman's reported error (these errors are from fully grown trees), RF is our forest implementation, DM is the Market with delta updates, and GM is the Market with Gaussian updates, and Speedup is the speedup factor of a depth 5 tree versus a depth 10 tree for evaluation. Bullets/daggers represent pairwise significantly better/worse than RF while +/-- represent significantly better/worse than RFB.}
\tiny
\centering
\begin{tabular}{|c|c|c|c|c|c|c|c|c|c|}
\hline
Data & $N_{\text{train}}$ & $N_{\text{test}}$ & $F$ & $Y$ & RFB & RF & DM & GM &\hspace{-2.mm}Speedup\hspace{-2.mm} \\
\hline
abalone & 4177 & -- & 8 & $[1.00,29.00]$ & 4.600 & 4.438 & 4.318\textbullet+ & 4.438 & 3.3 \\
friedman1 & 200 & 2000 & 10 & $[4.30,26.03]$ & 5.700 & 5.076+ & 4.701\textbullet+ & 4.429\textbullet+ & 1.8 \\
friedman2 & 200 & 2000 & 4 & $[-167.99,1633.87]$ & 19600.0 & 29343.562-- & 23200.438\textbullet-- &\hspace{-2.mm}21183.421\textbullet--\hspace{-2.mm} & 1.9 \\
friedman3 & 200 & 2000 & 4 & $[0.13,1.73]$ & 0.022 & 0.034-- & 0.029\textbullet-- & 0.028\textbullet-- & 2.0 \\
housing & 506 & -- & 13 & $[5.00,50.00]$ & 10.200 & 12.869-- & 12.056\textbullet-- & 11.947\textbullet-- & 2.2 \\
ozone & 330 & -- & 8 & $[1.00,38.00]$ & 16.300 & 16.976 & 16.964 & 16.932 & 2.1 \\
servo & 167 & -- & 4 & $[0.13,7.10]$ & 0.246 & 0.248 & 0.241 & 0.254 & 1.6 \\
auto-mpg & 392 & -- & 7 & $[9.00,46.60]$ & -- & 8.248 & 7.817\textbullet & 7.750\textbullet & 2.1 \\
auto-price & 159 & -- & 15 & $[5118.00,35056.00]$ & -- & 4699789.7 & 4524741.81 & 4431992.3 & 1.4 \\
breast cancer & 194 & -- & 32 & $[1.00,125.00]$ & -- & 1073.319 & 1071.820 & 1072.126 & 2.1 \\
diabetes & 43 & -- & 2 & $[3.00,6.60]$ & -- & 0.400 & 0.426\dag & 0.393 & 0.7 \\
forestfires & 517 & -- & 12 & $[0.00,1090.84]$ & -- & 4945.630 & 5445.001\dag & 5196.451\dag & 2.2 \\
machine & 209 & -- & 6 & $[6.00,1150.00]$ & -- & 3137.001 & 3127.932 & 2930.506 & 1.8 \\
triazines & 186 & -- & 60 & $[0.10,0.90]$ & -- & 0.016 & 0.015\textbullet & 0.015\textbullet & 2.0 \\
\hline
\end{tabular}

\end{table*}

\subsection{Fast Regression using Shallow Trees}
\noindent This experiment examined the aggregation capabilities of the regression market with shallow trees. In many problems, it is prohibitively expensive to train and even evaluate deep trees. In practice this is mitigated by enforcing a maximum tree depth. For example in~\citet{criminisi2010} and \citet{kinect2011} the regression trees were constrained to depth 7. However, this strict constraint on tree depth is prone to introduce leaves that do not generalize well due to prematurely halting tree growth. The specialized regression market of tree leaves can be used to weight the leaves. Poorly performing leaves will tend to have less weight thus improving the overall prediction accuracy.

\noindent In addition to the previously mentioned experiment details, regression trees were grown with a maximum depth of 10. Using the same depth 10 trees, MSE errors were computed for leaves no deeper than depth 5. Both depth 5 and depth 10 evaluations for training and test sets were recorded. The timings for the larger of the two sets were averaged over the 100 runs and used to compute the speedup. The markets were applied to the depth 5 leaves only. Since the market is just a linear aggregation of 100 leaves per instance, the reported speedup for forest is similar to the speedup of the market.

\noindent From~\ref{tbl:shallow} it can be seen that the depth 5 forest is roughly twice the speed of the depth 10 forest. On diabetes, the small data set, features and forest likely fit in cache giving the strange 0.7 speedup. DM performs significantly better than RF on seven data sets (in a pairwise set) while DM only performs significantly better on six data sets. However, DM performs significantly worse on two data sets while GM performs significantly worse on one. No method legitimately performs significantly better than RFB since RF is already better than RFB on those two data sets. The significantly worse results can be attributed to overfitting and/or poorly tuned reward kernel in the case of GM.

\section{Conclusion}
\noindent This work presented a generalization of the Artificial Prediction Markets from classification to regression with uncountably many outcomes. It introduced two types of update rules and demonstrated their learning ability through experiments on UCI and LIAAD datasets. Furthermore, it showed the capability of the regression market to aggregate shallow tree leaves into much better regressors than those obtained by voting. In future work we plan to use the market for regression with non-uniform noise levels and multi-modal conditional probabilities $p(y|{\bf x})$.


\bibliography{paper}

\end{document}